*Article*

# Reinforcement Learning with Probabilistic Boolean Networks Models of Smart Grid Devices

Pedro J. Rivera Torres [1,2,*], Carlos Gershenson García [1,3,4] and Samir Kanaan Izquierdo [2,5]

1 Centro de Ciencias de la Complejidad (C3), Universidad Nacional Autónoma de México, Circuito Centro Cultural S/N, Cd. Universitaria, Delegación Coyoacán, 04510, Ciudad de México, México.
2 Bioinformatics and Biomedical Signals Laboratory, Centre de Recerca en Enginyeria Biomèdica, Universitat Politècnica de Catalunya, Facultat de Matemàtiques i Estadística, Edifici U, C/Pau Gargallo, 5, 08028 Barcelona, Spain.
3 Instituto de Investigaciones en Matemáticas Aplicadas y en Sistemas, Universidad Nacional Autónoma de México, 04510, Ciudad de México, México.
4 Lakeside Labs GmbH, Klagenfurt am, Wörtherses, Austria.
5 Institut de Recerca Sant Joan de Déu, Esplugues de Llobregat, Barcelona, Spain
* Correspondence: pedro.rivera@c3.unam.mx.

**Abstract:** The area of Smart Power Grids needs to constantly improve its efficiency and resilience, to provide high quality electrical power, in a resistant grid, managing faults and avoiding failures. Achieving this requires high component reliability, adequate maintenance, and a studied failure occurrence. Correct system operation involves those activities, and novel methodologies to detect, classify, and isolate faults and failures, model and simulate processes with predictive algorithms and analytics (using data analysis and asset condition to plan and perform activities). We showcase the application of a complex-adaptive, self-organizing modeling method, Probabilistic Boolean Networks (PBN), as a way towards the understanding of the dynamics of smart grid devices, and to model and characterize their behavior. This work demonstrates that PBNs are is equivalent to the standard Reinforcement Learning Cycle, in which the agent/model has an interaction with its environment and receives feedback from it in the form of a reward signal. Different reward structures were created in order to characterize preferred behavior. This information can be used to guide the PBN to avoid fault conditions and failures.

## 1. Introduction

There is not a picture of the present that is complete without electrical power; it has become essential to our civilization. Electrical power has been a constant in our lives for almost two centuries since Faraday's discovery and the first alternating current power grid in 1886. Generating, transmitting and distributing electrical power has evolved from a commodity to a basic need during this time. This process has not changed much for a long time. Electricity is produced in many different ways, but the basic cycle is essentially the same: it is generated (via electromechanical generators, geothermal power, nuclear fission, solar and other means), then it is delivered to clients via a transmission-distribution network.

Most modern systems are still similar to the first ones, centralized, unidirectional electrical power transmission system with demand-driven control. In the last decades of the 20th century, local grids started to arise, and since the early 21st, the industry has attempted to take advantage of telecommunications improvements to solve the limitations imposed by centralization, and the challenges brought with the use of renewable sources and new technology. The European Union Commission Task Force on Smart Grids has defined these as an "electricity network that can cost efficiently integrate the behavior and actions of all users connected to it – generators, consumers and those that do both – in order to ensure economically efficient, sustainable power system with low losses and high levels of quality and security of supply and safety". Applying Signal Processing and Communications to the power grid has allowed a flow of data that is one of the defining elements of the smart grid. This includes the use of "Smart Devices", such as the Intelligent Power Router (IPR). This device [1] was inspired on Internet routers, and it has a degree of intelligence that allows it to switch lines and shed loads. Devices such as the former allow the Electrical Power Distribution System (EPDS) to become reliable, resilient, flexible and efficient. With them, decisions can be made in the event of power failures or component malfunctions, coordinating with other devices in their vicinity to react to load, demands, faults, and emergencies. The basic elements of an IPR are show in Figure 1.

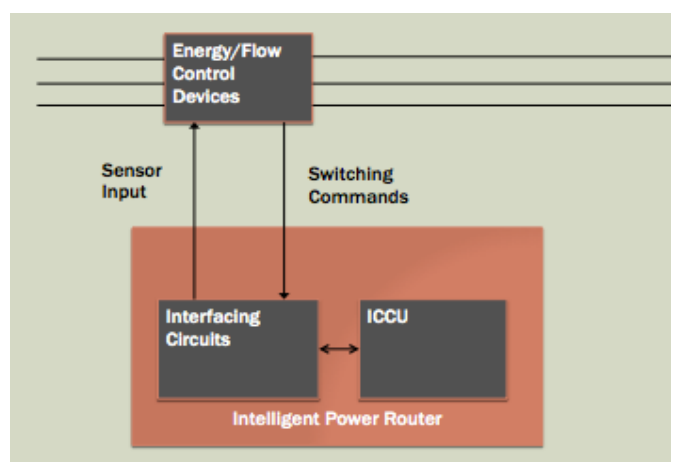

**Figure 1.** Basic Elements of an IPR

EPDS that have incorporated IPRs (EPDS-IPR) have also the capacity of automatic service restoration, if a network of IPRs is deployed strategically throughout the power grid, and if they are programmed for the exchange of information to manage and reconfigure the network following a rule set, any time a perturbation occurs. This allows survivability, and better use of system resources.

Designing these EPDS-IPR networks is a very complex task. There is no specific model that can guide the designer. The devices have to be configured with pre-set instructions on how to react when a particular set of conditions have occurred. Another challenging task is to make these grids adaptive [2], and not just follow a hard-wired set of instructions. A much more favorable situation is that the network can act autonomously and self-reconfigure in the event of a perturbation, i.e., loss of a power source, higher demand in critical loads, or sabotage. IPR devices when interconnected can be modeled as an intelligent Probabilistic Boolean Network, which is a complex-adaptive system that can learn from its steady state behavior, adapt, with self- organization and resilience. Methodologies for modelling based on a Probabilistic Boolean Network (PBN) have been presented in [3-12], validating the use of PBNs as a modelling mechanism for industrial processes and Smart Grids using IPRs, enabling the simulation of several scenarios. This has the potential to allow designers to better program the devices and design a more robust network. We would like to imbue EPDS-IPRs with the intelligence that allows them to survive a wide set of perturbation events that are practically impossible to predict.

Biomimetic approaches have been used to analyze and solve complex problems in general and for EPDS design in particular. Frameworks that are qualitative in nature, such as PBNs, permit the description of biological system networks, with no property losses that are relevant to the system, and allow the representation of complex-adaptive behavior, such as self-healing, and self-organization. Probabilistic Boolean Networks are used in bioinformatics for Gene Regulatory Networks (GRN) modeling. GRNs are DNA segments in a cell that influence other segments and substances in it indirectly, to rule the level of expression of a gene or a set of genes. They are utilized to master and perceive the main rules that command the regulation of genes in genomic DNA. These PBNs are state-transition systems satisfying the Markov Property; they have no memory, so they are not reliant on previous states of the system). Proposed by I. Shmulevich and E. Dougherty [12] through extension of Stuart Kauffman's N-K or Boolean Network (BN) concept [13,14], they mix the rule-based modelling richness of BN and introduce probabilistic behavior. These PBNs are built upon a collection of constituent BNs which are assigned selection weights or probabilities, in which every BN can be considered a "context". Information for every one of the cells comes from antithetical sources; each represents a cell context. For every point in time $t$, a particular system can be commanded by a single BN, and the PBN will change to another context, or constituent BN at a different time, based on a particular switching probability. The methodology for using PBNs in manufacturing engineering systems was proposed in [12] and [8], with continued development in [3-7, 9-11].

In genomic research, the main focus is to discern the manner in which cells exercise control and perform extensive numbers of operations that are needed for their operation and function. They are massively parallel and highly cohesive systems, and a path that considers a perspective supreme to that of a single gene is needed so we can understand these biological processes better. Genes, cells and molecules are networked systems that require a deeper understanding, in order to manufacture improved medicines and delivery mechanisms for treating and eradicating human disease. A mechanism for treating and processing massive quantities of data using computational methods and model checking can be used to understand the rules that govern them, and make more accurate predictions about how these systems behave. EPDSs are akin to GRNs because, in order to understand the main rules that control them, and to make accurate forecasts on how they will behave, endure or decline under a collection of prospects, models that correctly describe the system and its behavior are essential. The harmonized synergy, interaction and governance between genes and their products form these chains, in which gene expression is an important factor.

In this research, the use of Probabilistic Boolean Networks, already applied auspiciously in manufacturing engineering systems, will be broadened to permit analyzing IPR reliability and trust, and the scrutiny of faults that lead to catastrophe. As our main contribution, we explored the PBN's model capacity for performing a basic Reinforcement Learning (RL) cycle, and we explored RL as a means for directing the network's evolution in order to increase the network's resilience, working towards achieving automatic learning and control of itself using RL.

## 2. Preliminaries and Theoretical Background

A review of Boolean and Probabilistic Boolean Networks, Reinforcement Learning, and a basic understanding of Electrical Power Distribution Systems, and Intelligent Power Routers is presented in the subsections.

### 2.1. *Probabilistic Boolean Networks in System Modeling and Simulation*

Kauffman N-K or Boolean Networks (BN) [13-14] and PBNs [15-18] have been studied for biological systems modeling and their dynamics, and to infer their behaviors with statistical data analysis and simulation. This application is very well documented in bioinformatics for biological systems modeling [19-23], and for GRNs description [24-29]. The mechanism of intervention [17] is used to conduct the evolution of the PBN away from unfavorable conditions or states, as are those associated with illness.

Kauffman's BNs are a finite grouping of Boolean nodes [30, 31], in which states quantize to 0 or 1, (although in PBNs, alternative quantizations are possible), for which, state is decided by the present state of other nodes/genes in the network. The set of entry/input nodes in a BN are known as regulatory nodes, with a collection of Boolean functions (known as predictors), that dictate the future values of the different nodes. When the set of genes and their respective predictors are defined, the network is defined as well. PBNs are, in essence, a tree of BNs for which, at any particular time period, the node state vector transitions are established by one of the rules of the constituent BNs. Formally, a Kauffman Network is a graph G(V, F) defined by the set

$$V = \{x_1, x_2, \dots, x_n\} \quad (1)$$

that contains all of the network's nodes, and the set

$$f_i^{(j)}(i = 1, 2, \dots, l(j)) \quad (2)$$

instead of the single predictor per node in a BN, that can be selected to determine the future state of node $x_j$. The probability of selecting $f_i^{(j)}$ as the predictor for the node is given by $c_i^{(j)}$, where

$$0 < c_i^{(j)} \leq 1, \sum_{i=1}^{l(j)} c_i^{(j)} = 1, \forall j = 1, 2, \dots, n \quad (3)$$

A useful metaphor is to think of the PBN as a tree of BNs, and each BN is selected with a particular probability.

Let $f_i$ denote the *i*th possible realization of the network, with

$$f_i = \left(f_{i_1}^{(1)}, f_{i_2}^{(2)}, \dots, f_{i_n}^{(n)}\right) \quad (8)$$

for every

$$1 \leq i_j \leq l(j), y\, j = 1, 2, \dots, n \quad (9)$$

A realization of a PBN is one of its constituent BNs. The maximum number of realizations is given by

$$D = \prod_{j=1}^{n} l(j) \quad (10)$$

In [8], the authors validated that PBNs are appropriate for modeling engineering systems through a system model, that was verified using model checking, and the simulation results compared with real machine data. In [7], this methodology was applied to a manufacturing process, to gather quantitative occurrence data for DFMEA. In [6], the methods were further expanded include the application of PBNs in industrial manufacturing processes, using intervention (guided perturbations) as guide to move a system away from fault conditions and catastrophe, thus postponing its failure. A formal and thorough description of BN and PBN is presented in [18].

### 2.2 Reinforcement Learning

Born in the field of Behavior Psychology, Reinforcement Learning (RL) [32-33] is considered an area of Machine Learning (which can be defined as the design and analysis of algorithms that can improve on the base of experience) in the field of Computer Science, and it is concerned with how agents should perform actions in a given environment such that they maximize a cumulative reward signal. In Reinforcement Learning, a learner, or agent, is not told what to do or which set of actions to take, rather it must discover the group of actions that achieve an optimal reward by trying them. The use of trial-and- error and delayed rewards are the two characteristic features of this approach. RL is studied in many other disciplines, such as control theory, operations research, statistics, and game theory. RL allows the software agent to learn a correct behavior based only on feedback from the environment, automating the learning scheme and extinguishing the need for human expertise, cutting the time needed to devise a solution. There are multiple solutions to a RL problem, but the most common approach is to allow the agent to select actions that yield a maximum reward in the long run, by using algorithms with infinite horizon. This is performed in practical terms by learning how to estimate the expected future rewards of states. The estimates are adjusted through time by propagating part of the future state's reward, and if all states and all actions are tried numerous times, an optimal policy can be learned.

A Reinforcement Learning agent learns by interacting with its environment. The RL agent acquires knowledge from the result of its interactions with the environment, instead of being taught explicitly, and selects its actions based on past interactions (called exploitation), or by making new choices (exploration). The reinforcement signal (mostly numerical in nature) it receives is a reward that encodes the success (or failure) of a given action's outcome, and it seeks to acquire knowledge by selecting actions that maximize the cumulative reward over time. Figure 2 illustrates the standard Reinforcement Learning cycle.

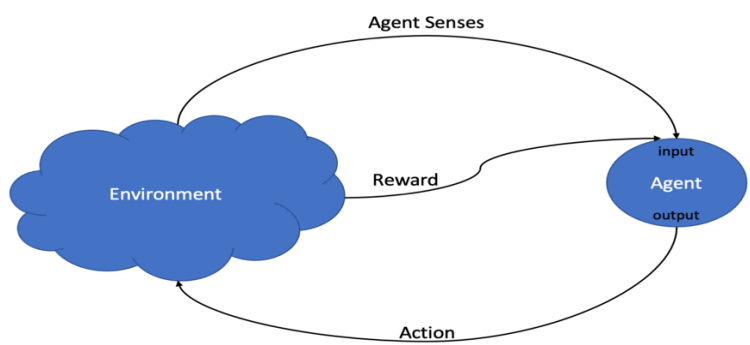

**Figure 2**. Standard Reinforcement Learning Cycle

In a standard Reinforcement Learning model, the learner is known as the agent, who makes decisions and is connected to its environment through perception and action. Agent and environment interact at a sequence of steps in time, t, and at each interaction step our agent senses the environment for information and determines the state of its "world". Based on this information, the agent chooses and takes an action. The information of the state of the environment constitutes the input of our agent, and the action chosen by our agent becomes its output. The actions taken by the agent in each step change the state of its environment, and its own state. A time step later, the state transition's value following the action taken is given to our agent by its environment as a numerical value, called reward.

Reinforcement learning differs from supervised learning, another form of learning studied in machine learning where the agent learns from examples that are provided by a supervisor, which is external to it. A challenge that is present in RL and not in other learning methods is that we have to choose and/or balance exploration and exploitation. An agent that uses exploration discovers and tries new actions to see if they produce more or less reward, but an agent that uses exploitation uses preferred and tried actions that in the past have been successful at producing reward. It also considers the whole problem, in uncertain environments the agent doesn't consider sub-problems and sees how everything fits into the whole picture, starting with an agent that is complete, interactive and with explicit goals, sensing aspects of its environment and choosing actions that influence it.

2.2.1 Supervised and Unsupervised Learning

A representative set of pairs of states and actions is provided by a teacher to the agent in supervised learning. The agent has to modify its strategy for selecting actions so that its actions get closer every time to the selected target actions. Therefore, the main problem in supervised learning is then the approximation of a functional mapping from states to actions that is an unknown to the agent and known to the teacher, which can be done with neural networks, fuzzy systems, or other learning models. This is impractical for complex problems, because of the inability to specify a representative set of pairs of states and actions, making finding optimal solutions unknown in some instances.

An agent that performs unsupervised learning in its purest form is perceiving the states of the process it has under control, but will not get any information about the actions, and the control strategy is not evaluated. Unsupervised learning cannot be used to learn control strategies. A typical application of it is the identification of structure in data, as in data clustering.

2.2.2 Reinforcement Learning versus purely Unsupervised Learning

Just as in unsupervised learning, our agent performs reinforcement learning and receives no information about an optimal strategy for control, but in RL the agent gets rewards, or reinforcement signals that provide feedback about its control strategy. With these signals, the agent can improve the strategy, giving intelligence to the trial-and-error process.

The problem faced by our agent in RL is that it has to learn its behavior through trial-and-error interactions with its environment. Two main strategies are used for RL problem solving: an agent can choose to search in the behavior space to find a behavior that is appropriate to its environment (the approach used in genetic algorithms and programming), or it can use statistics and dynamic programming to estimate a utility function of taking actions in states of its environment.

2.2.3 Main components of a Reinforcement Learning system

In addition to the agent and the environment, the principal components of a Reinforcement Learning system are:
- The **policy** dictates the way that our agent will behave at any given time. It maps states of the environment to a set of actions that are going to be taken when the states are reached. This policy is central to our agent, since it is the only thing needed to determine the agent's behavior.
- The **reward function** is the reinforcement signal, and in our RL problem, defines the goal of the agent, by mapping the perceived state or states of our world to a numerical value. This way we know which state is more desirable. In RL, our agent's only purpose in "life" is to maximize the total reward it will receive, and our agent must choose then which actions contribute to that goal. Reward functions may be stochastic, and are the basis for changing the policy of the agent. A strong assumption of the RL framework is that the reward signal can be unequivocally and directly observed, as the feedback the framework receives is part of the environment in which the agent is working on.
- The **value function** estimates the expected cumulative reward of a given state. This way, it specifies what is better for our agent in the long run; whether or not a given state is desirable considering the states that follow this one and the rewards available in them. They are secondary to rewards and serve as predictions of them. Action choices are based on values. Values are much harder to determine than rewards. Rewards are essentially given by the environment to the agent, whereas values are estimated and then re-estimated from the observations that an agent makes over time.

- The **model** of the environment is a simulation of the behavior of the problem's environment. It is required by some RL algorithms, although in practice it is only used in relatively simple problems.

2.2.4 Markov Decision Processes

Reinforcement Learning problems are well modeled as Markov Decision Processes, or MDPs. Named after Russian scientist Andrey Markov, MDPs can be viewed as RL tasks that satisfy the Markov Property. When a stochastic process satisfies the Markov Property, it has a memoryless property, or the conditional probability distribution of its future states is only dependent on the present state and not the previous sequence of events. MDPs are discrete time stochastic control processes that are useful for studying and solving optimization problems through Dynamic Programming and RL. MDPs consist of the following:

- A set of states, **S**. The states are the inputs to our learning system. They represent all the information necessary to perform optimally.
- A set of actions, **A**.
- A reward function. $R : S \times A \to \mathbb{R}$ Our learning system will execute an action in each state, and each action causes a transition. Because of the Markov Property, rewards are only dependent on the current and the successor state, and do not depend on past information.
- State transition function $T : S \times A \to \Pi(S) \mid \Pi(S)$ is a probability distribution over S.
- Policies are sequences of mappings in the form $\Pi : \{\pi_0, \pi_1, \ldots\}$, where $\pi_k$ maps the state $s_k \in \mathbf{S}$ to an action $a_k = \pi_k(s_k) \in \mathbf{A}(s_k)$. When both state and action spaces are finite, MDPs are said to be finite.

The Value Function, $V^\pi(s)$, in Reinforcement Learning is of extreme importance. It estimates the expected cumulative reward of state *s*. In MDPs, the value function can be defined as

$$V^\pi(s) = E_\pi\{R_t \mid s_t = s\} = E_\pi\{\sum_{k=0}^{\infty} \gamma^k r_{t+k+1} \mid s_t = s\} \tag{10}$$

where $E_\pi \{ \}$ defines the expected value when the agent follows the policy π. The terminal state's value, if it exists, is zero. $V^\pi$ is the state-value function for π. Most RL algorithms estimate value functions. A value function is a mapping of states that provide an estimate of how fit is it for the agent to be in a given state, defined in terms of the future rewards to be expected, or the expected return. They are defined with respect to a given policy. We also define $Q^\pi(s, a)$, the action-value function for π, as:

$$Q^\pi(s, a) = E_\pi\{R_t \mid s_t = s, a_t = a\} = E_\pi\{\sum_{k=0}^{\infty} \gamma^k r_{t+k+1} \mid s_t = s, a_t = a\} \tag{11}$$

In many RL algorithms, the action-value function *Q* is used instead of the value function *V* because it easily lets the agent choose the action with higher expected rewards.

2.2.5 The Reinforcement Learning Problem

The RL agent interacts with its world in a series of time steps. With each discrete time step *t*, the agent receives an observation $o_t$ and a reward $r_t$. An action at is chosen from the group of actions available to the agent, and executed within the environment. This moves the environment to a new state $s_{t+1}$. A new reward for the transition and new state is now determined. The agent needs to accumulate as much rewards as possible.

The RL problem is defined as finding a policy for the agent, a mapping for the agent that will specify the action that the agent will take when in a given state. Once an MDP combines with a policy as such, the action for each state is fixed and behaves like a Markov Chain. RL is not considered a technique for the solution, but rather a way of formulating a problem [34]. In [35], the basic problem that is apt for the use of RL is formulated, where a system needs to interact with the environment in order to achieve a certain goal and based on the current state's feedback, what action should it perform next? RL in itself is the way of learning the correct action to be taken in a given situation based solely on feedback obtained from the environment [32]. For our purposes, feedback is a numerical reward that we assign to the actions that an agent takes. RL agents can learn off-line or online. Off-line learning is similar to the knowledge acquired by a student from a teacher; the agent is taught what is needed to know before venturing into the environment. Online is more spontaneous; similar to the way a child learns how to walk, where knowledge is acquired in real-time. The agent explores its environment and it is constantly adding experiences in order to make better future decisions.

*2.3   Electrical Power Distribution Systems and Intelligent Power Routers*

2.3.1 Electrical Power Distribution Systems

Electrical power generated at and transmitted from generation plants is the product of the transformation several energy sources (coal, natural gas, petroleum, nuclear, geothermal, solar, tidal, etc.) into electrical current [36]. Managing this vital resource is guided by the necessity to secure a stable and persistent supply of energy, in spite of demand fluctuations. Electrical power plants have grown in capacity and size since they were first built over a hundred and fifty years ago. The generation of electrical power generally occurs distant from where it is ultimately consumed. Therefore, consumers are usually separated from electrical power plants by great distances.

Two distinguishable types of networks that interconnect consumers to generation sites exist. These types are:
- Transmission networks: covering large areas, they make sure that most or all regions of a country are covered and provide the service. These high Voltages (in the range of 230 kV or 138 kV) allow for the minimization of losses in its transmission. Different lines are bundled together at electrical power substations, and the networks eventually interconnect and feed power to distribution networks, that reach end-customers.
- Distribution networks: these are engineered to provide power to smaller areas and have lower voltages than transmission grids, but are denser, because they are meant to serve electricity to the final customers. Lower voltages are used for safety, and security reasons, and due to installation costs. They are also able to provide several voltage levels to different end users, through transformers.

Industrial, commercial and residential end users have to receive reliable electrical power at their facilities or homes. Several factors, natural and artificial, in the process of generating, transporting and distributing electricity can damage equipment (wind, ice, storms (thunderstorms, typhoons, hurricanes), vegetation growth that can induce short circuits, and other nature-induced or human disasters, as well as malicious perturbations). Some factors cannot be predicted and have to be taken care of in real time. Other events and factors can cause to network state unbalance. As an example, variations in temperature may cause changes in electrical loads, and overall demand for electrical power varies with time, season, weather, etc. Some of these factors affect supplied power quality, while other factors cause emergency situations that force network operators to disconnect power to regions that cause problems, in order to prevent chain reactions. Other severe situations may cause power outages or network power imbalance. Intentional power outages have to be limited and minimized.

Electrical Power Grids are almost always managed from control rooms. Some can be telemetered, such that control engineers have accurate real-time information about their status. They can also have protection equipment that can be actioned from within the control room, so larger failures are prevented. There are instances in which telemetry may be cost ineffective, and aberrant network states can be reported by operators, engineers, workers, or customer communication. System repairs and maintenance may be performed manually by skilled workers.

Electrical substations [36] can have several busbars, and two of these may be interconnected via a switch or a conductor line. Both extremes of the power line are connected to a breaker. A breaker is a standard protection mechanism that has a relay, that can automatically open in case of a short circuit, giving it the ability to disconnect all or a single circuit from the remaining network. Messages with alarms to control rooms can be generated as well, and with these, engineers can have the ability to control the state of the breakers. The main objective in fault management of an EPDS is to restore the power supply quickly to as many end-users as possible. Since in an EPDS there may be different routes through which power can be served, the EPDS can be switched to select alternative routes through breakers and switches that can bypass the areas, lines, or devices that cause problems. The need arises to isolate and determine any malfunction in protective equipment, generate correct diagnoses from the alarm messages that are received, and continue to postulate a plan to safely and efficiently restore electricity to the largest number of end-users.

EPDS [36] are a group of sources and power lines that operate under common supervision in order to provide electrical power to end-users. Systems for electrical power delivery are formed by joining Distribution and Transmission Systems. Transmission Networks are meant to transport high voltage electricity over longer distances. Their high voltage loads are reduced at major load centers and then distributed to customers, where distribution Networks transport electricity from the Transmission Network to the customers. EPDSs are ubiquitous, from large ships to modern data centers. For our scope, we consider only Generation and Transmission Systems.

2.3.2 Intelligent Power Routers

IPRs [1] are the principal components of a Smart-Grid that was developed as a distributed architecture for decentralized coordination, control and communication between power system components. Intelligent control and planning of network operations is built into smart computing devices attached to sources, power lines and other power network devices, thus allowing them to have a picture of actual network conditions, assign resources to respond to failures, priority or demand, etc. They are configured on a Peer-to-Peer (P2P) network architecture, and in the event of a failure, they make local decisions and coordinate with other devices in their neighborhood to return the system into operation from an undesired state.

Currently, the control of electrical power generation and distribution, even if redundant generators and lines are present, is done in a centralized way. Future EPDS should be capable of distributing coordination and control of generation and distribution tasks throughout the network when contingencies or emergencies arise. IPRs were engineered for survivability, fault tolerance, scalability, cost-effectiveness and continuous unattended operation. At its core, the IPR is a power flow controller with embedded software. An IPR has two principal components: Interfacing Circuits (ICKT) and an Integrated Control and Communications Unit (ICCU). The ICKTs operate power flow control and sensing devices, such as breakers, capacitors, and transformers. They can also receive network status information from sensors and dynamic system monitors. They have direct control of the ICCU, and with their logic and software calculate how to route power, change loads and take any corrective or preventive actions that enhance safety, stability and security. The network architecture and communication protocols are similar to the Internet Protocol (IP) Local Area Networks. A load connected to an IPR can be assigned a priority, and contrary to non-smart power networks, when a power source fails, the ICCU of an IPR reacts to this failure through reconfiguration of the network, so that the load with the highest priority may be served.

**3. Materials and Methods**

With the following methodology, faults and failures can be categorized for a single IPR's failure modes in an EPDS. We propose establishing the model using PRISM [37], to verify its use and formal correctness using Probabilistic Computational Tree Logic (PCTL). The models were built in PRISM by constructing three modules: one for the environment in which the device operates, a module for the IPR's Probabilistic Boolean Network, and a reward structure. The actual state of the device PBN's nodes is in the second module, which uses the state of the variables available in the environment module, and applies the corresponding Boolean Predictor Functions to transition to the next state. With the values of these variables as a base, and the device's failure modes, the state of the IPR variables is changed, giving us the device's current state. In this manner, given the device's failure modes (which are based on the possible failure modes of its components), the model produces the failure modes corresponding to the system as an output.

To calculate individual IPR reliability, we have divided them into three principal subsystems: power hardware (power circuit breakers), computer hardware (used for IPR-to-IPR communications, routing and CPU functions), and the software that manages the device. The reliability estimates of each of the subsystems that compose the IPR are provided in [1]. The reliability of a circuit breaker was obtained from data sheets as 0.99330. Each IPR has two circuit breakers, a main breaker, and a redundant secondary. The reliability of data routers is estimated at 0.9009 (in a year). Lastly, software reliability is estimated at 0.99.

PBNs can precisely emulate an EPDS with IPRs since this has coincidences with GRNs that have been modeled with BNs and PBNs. As a first step, the PBN representing the EPDS is built. Each modeled component of the EPDS is equivalent to a gene (node) in a GRN, where a gene can assume one of two states; 0 means the IPR is ON and 1 meaning it is OFF (by the convention established in [1]). For each node, the Boolean functions that determine the state of its IPR in time $t + 1$ are applied, given the state of the EPDS's nodes in time $t$.

In the next step, a matrix for every node is built, to construct its Predictor Function. When calculating the predictors, only relevant nodes, those affecting directly the status or state of the node under study, are considered. All nodes that do not directly affect the current node's state are ignored. As per the connections between relevant nodes and the observed node, the equation or set of equations (constructed with the basic Boolean operators) that determine the state of each node are presented. For every node, there exists a set of equations (one or more per node). These Boolean Functions are resolved from the examination of the relationships between each node and its relevant nodes. All possible states of all relevant nodes are analyzed, and an evaluation is made about the next state of the node in time $t+1$, given the state of all relevant nodes in time $t$. This method proposed adapts the Fault Detection and Isolation (FDI) scheme described in [38], and shown in Figure 3, where a model is used for describing the normal operation of the process and another model is used to describe each of the faults or failure modes.

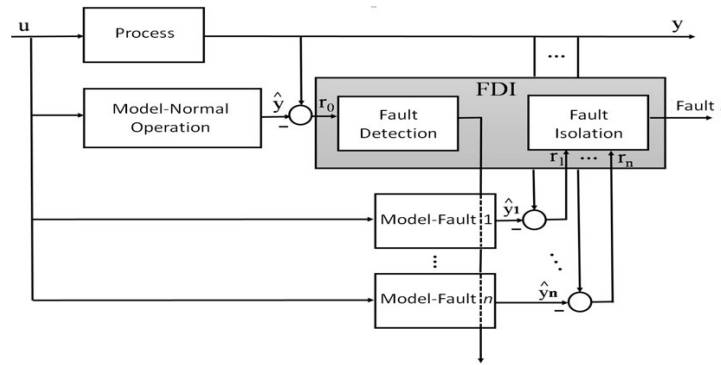

**Figure 3.** Fault Detection and Isolation (FDI) method from [38].

PBNs possess a characteristic called self-organization, and they do so into attractor states [31]. Attractors are cyclic states that, in the case of the models under study, are related to the failure modes that the system exhibits. There are similarities between the construction and semantics of the models we present and those in [7]. By characterizing the failure modes of the device under study, the models can, with model checking, characterize the state of their nodes to determine the faults and failures correlated to the device's fault conditions. This methodology is flexible, and the design of the network model and its state transitions depends on how much resolution the experts need, based on design specifications. Complexity and expression are scalable in this method, depending on the needs of the experts.

Device operation has been modeled by simulation of the network's components, taking into consideration the reliability analysis in [1]. These simulations were performed for the IPR by modeling of its relevant components, based on the data of their Mean Time Between Failures (MTBF). The model is able to detect and isolate single and multiple IPR failure modes.

### 3.1. Classification

We begin by classifying the different states of the IPR's components, in order to properly create models of the device. Following the methodology in [38], we begin describing the system's failure modes, starting from its normal operation and continuing into modeling the different types of fault in the system.

Each subsystem is perceived as being in one of two different states:

Breakers
- 0 – the breakers are able to close/switch properly
- 1 – the breakers do not close/switch properly

Router
- 0 – the data router communicates/sends-receives information in the network properly
- 1 – the data router does not communicate/send-receive information in the network

Software
- 0 – the software makes correct decisions
- 1 – the software makes incorrect decisions

The state of the device is, therefore, a set of states of its subsystems. There is redundancy in the breakers, because, since all configurations of the IPR break into a series system, and the reliability of a series system is below the reliability of its lowest component. Therefore, the only way to increase the reliability of the IPR would be to provide a redundant path to the breaker. The device can be in 16 states that go from all subsystems operational (0 0 0 0) to all subsystems in failure (1 1 1 1). Some of these states, such as the failure of a single breaker, are identical, and after merging, there are 12 unique states. Table 1 summarizes the Categories, Types, and states that constitute these categories in the IPR.

**Table 1**. IPR Failure mode classification and their corresponding states [1].

| Category | 1 | 2 | 3 | 4 |
|---|---|---|---|---|
| Type | Fault | Normal Operation | Failure | Fault |

| Description | On Active Signal (AS, Switching Event), the IPR works as intended. On Inactive Signal (IS, non-switching event), IPR does not work as intended. | On AS, the IPR works as intended. On IS, the IPR works as intended. | On AS, the IPR does not work as intended. On IS, the IPR does not work as intended. | On AS, the IPR does not work as intended. On IS, the IPR works as intended. |
|---|---|---|---|---|
| States | S9 | S0 | S3, S4, S10 | S1-2, S5-8, S11 |

Failure probability for each component is assumed to be independent of each other. Reliability estimates for each of the device's components was detailed in Section 3.

The relevant genes of the IPR's PBN are its data router, software, and the main and secondary breakers [36]. For these, the state of their components determines the failure mode they are currently on, as per the categories. Category 1 is a type of fault, where the IPR acts appropriately and changes the state of the breakers on an Active Signal (AS), but may change them when switching is unnecessary. Category 2 describes the normal operation mode of the IPR. Category 3 describes a failure (catastrophic) of this device. Lastly, Category 4 describes a fault condition on which the device does not act upon an AS, and may also switch the breakers unnecessarily when there is no AS. Table 2 presents the predictor Boolean functions for every IPRs subsystems, based on their configuration.

**Table 2**: Predictors and Selection Probability, IPR PBN

| Component | Predictor | Selection Probability $c_j^{(i)}$ |
|---|---|---|
| $x_1$, Software | $x_1(t+1) = x_1(t)$ | 1 |
| $x_2$, Router | $x_2(t+1) = x_1(t)\ \&\ x_2(t)$ | 0.9611 |
| | $x_2(t+1) = x_1(t)\ |\ x_2(t)$ | 0.0389 |
| $x_3$, Main Breaker | $x_3(t+1) = x_1(t)\ \&\ x_3(t)$ | 0.9611 |
| | $x_3(t+1) = x_1(t)\ |\ x_3(t)$ | 0.0389 |
| $x_4$, Secondary Breaker | $x_4(t+1) = x_1(t)\ \&\ x_4(t)$ | 0.9611 |
| | $x_4(t+1) = x_1(t)\ |\ x_4(t)$ | 0.0389 |

This permits the prognosis of fault conditions; those that do not cause a total failure, but rather failure modes that will lead to instances where the device continues its operation, but does not perform the required task to specifications. These are unhealthy states of the device, and they should be treated, or they will otherwise lead to failure. For the device under study, the failure modes described in [1] were used, and an expert determination was made as to which device components and failure modes produce a failure or a fault.

PRISM's Property verification in PCTL was used to determine the maximum probability of occurrence of the failure modes that could evolve to a fault or a failure. From an initial state for the IPR, such as Category 2, a determination is made about the maximum probability of reaching one of the different identified failure modes. Property verification in PRISM permits the verification the models, and they also permit, through experiments, to reach an estimate regarding when in time a fault is certain.

## 4   Results

PRISM was used to validate the model quantitatively. These experiments were performed using a PBN representation of the IPR. Its main components (router, software, and breakers) are modeled, and their interrelationships expressed as Boolean Functions, or predictors. These components are considered the PBN's nodes, which give as output the overall state of the device. In the aforementioned experiments, time is expressed in hours (h). The rewards structure assigns a reward to the interaction of the PBN agent with its environment. A reward of '1' has been assigned to the state in which all components of the IPR are operating correctly. In this way, the agent is able to obtain a feedback signal, based on its actions. The main objective of the PBN-RL agent is to remain in a normal operating state through its operation.

We performed reward-based property experiments to test the model's capacity to emulate the standard RL cycle. We studied the agents combined actions in the environment, and we also studied the actions individually. The experiments assess the model's capacity to perform the standard Reinforcement Learning Cycle. In the standard RL cycle, an agent interacts with its environment and receives feedback upon performing actions in the form of a reward or cost. A PBN-based model was established in PRISM using an MDP, with a module for the environment, and a module for the PBN, with its predictors. The actions of the model correspond to the different states in which the model can be, which are correlated to the classifications previously presented. These classifications correspond to the device's different failure modes, as per the reliability analysis. In this scheme, the only missing element would be to assign a reward for the actions that are to be reinforced, so that the agent can receive feedback from its environment. PRISM has a rewards structure that can be used within the model's specification to assign rewards to states or sets of states. Currently, all assigned rewards in PRISM have to be positive, and therefore we cannot assign penalties or costs to states or sets of states. It is possible, however, to create multiple reward structures within the same model. With these multiple reward structures, we can analyze the effect of the different actions in the model. We are also able to conduct reward-based experiments that can provide information about the maximum cumulative reward over a period of time, for a particular action or state. In this model, the different failure modes of the system have been assigned a reward structure, where the corresponding states of each mode receive a '1' as a reward. We've studied the effect of these rewards separately, because although we value the benefits that using model checking provides, we are unable to, with this tool, assess the effect of assigning costs and rewards at the same time. We understand that the current benefits of the use of model checking outweigh its limitations.

The first experiment conducted was performed to determine the maximum expected reward for the agent interacting in its environment and executing any of its actions. The rewards structure assigns a reward of '5' to the normal operation mode, a reward of '1' to any of the fault modes, and a reward of '0' to the failure of the IPR (a higher reward for the action that we would like the device to reinforce more). This was executed through verification of the following property:

"*Rmax =? [C<=time]*"

Figure 4 presents the results of this experiment to assess the maximum expected reward of the agent when interacting with its environment in the standard RL cycle.

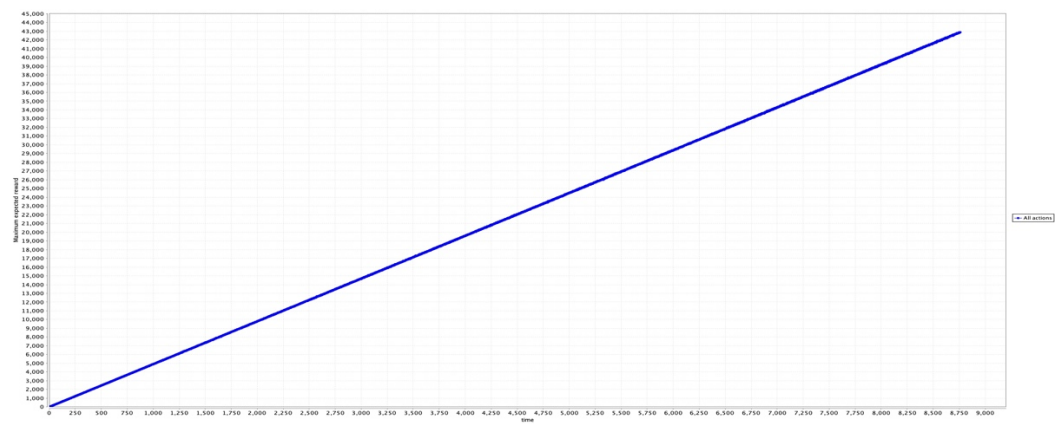

**Figure 4**. Maximum Expected Reward for the RL agent interacting with the environment

In this experiment, the final value of the reward is 42909. The agent performs its actions and receives a feedback in the form of a reward from each action from the environment, resulting in the plot from Figure 4. To recall, Figure 5 presents a graph of the maximum probability of occurrence of the normal operation action. Since this property reaches 100% probability quickly, a year of operation is not plotted.

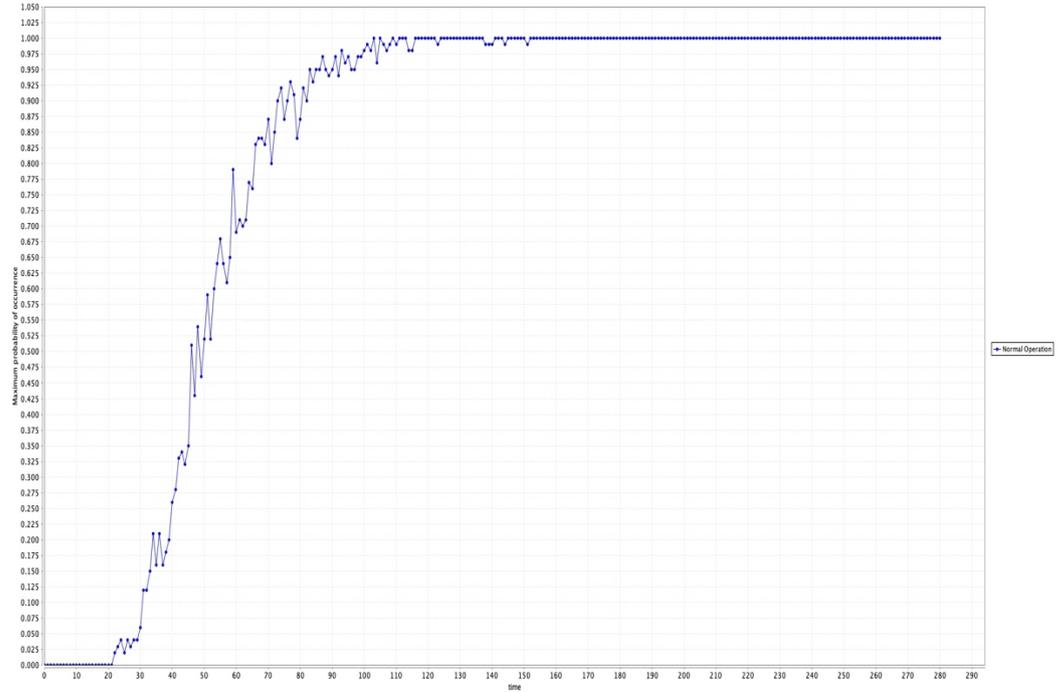

**Figure 5.** Maximum probability of occurrence of the Normal Operation Mode

Figure 5 approximates the cumulative distribution function of a bathtub curve used in reliability engineering. Figure 6 presents the maximum expected reward for the normal operation action. It shows a plot of the maximum reward obtained for the normal operation mode of the IPR in time, over a period of one year of operation. The final value of the maximum cumulative reward for this action is 8620, which translates into the device receiving a reward of '1' for every hour of normal operation, or 8,620 hours of normal device operation in the simulation.

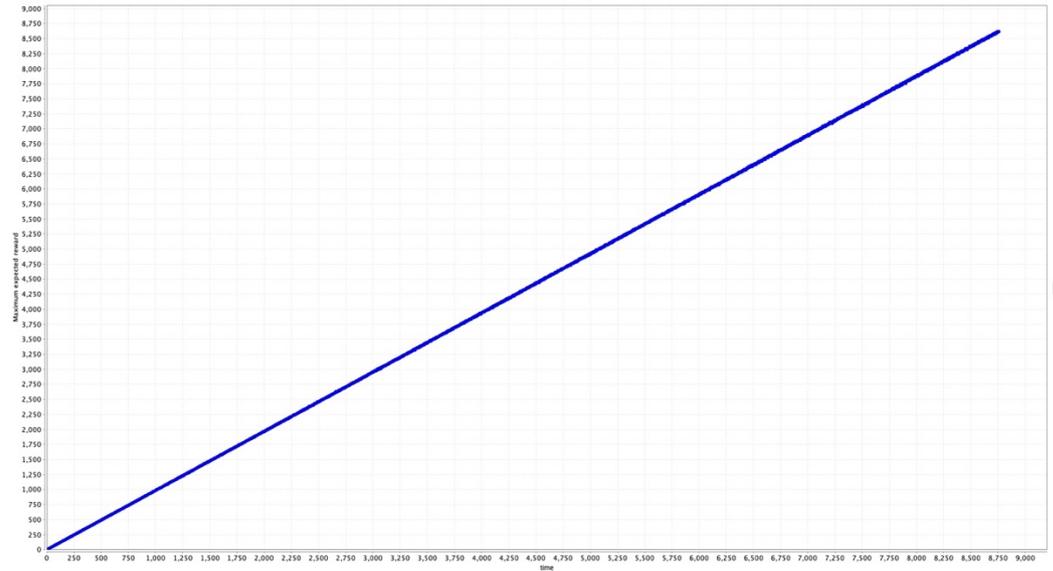

**Figure 6.** Maximum Reward obtained for the Normal Operation mode of the IPR in one year of operation

This is the action with the largest cumulative reward, as the set of states that are part of this classification have the highest probability of occurrence. The rewards for all of the other actions presented have a smaller cumulative reward, as these actions also have a lower probability of occurrence. We can adjust the scale of the Maximum Occurrence experiment in Figure 6 to match the time axis to Figure 5. Figure 7 shows this adjustment.

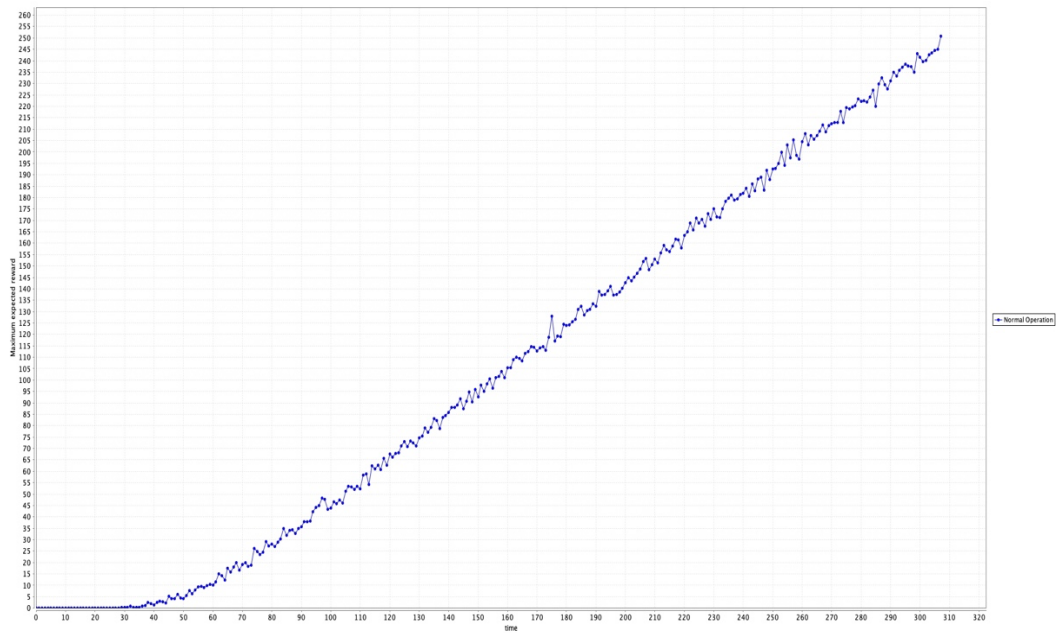

**Figure 7.** Maximum Reward for Normal Operation, adjusted scale

The reward initially is low in the very early hours of operation, corresponding to the early failures (or infant mortality) period, and as the device enters a steady state, its expected increases linear reward rate increases almost linearly.

As PRISM supports multiple reward structures within a single model, each reward structure needs to be identified when running an experiment, as

"R{"normop"}=? [C <= time]"

where the property used the "normop" (for normal operation) reward structure of the model. The rest of the experiments were executed with similar properties.

Figure 8 presents a plot with the results of a Maximum Cumulative reward for the Failure of the IPR over a year of operation. The final value for the maximum cumulative reward is 78, which reflects a total of 78 hours for a period of one year in which the device was in a catastrophic failure in the simulation.

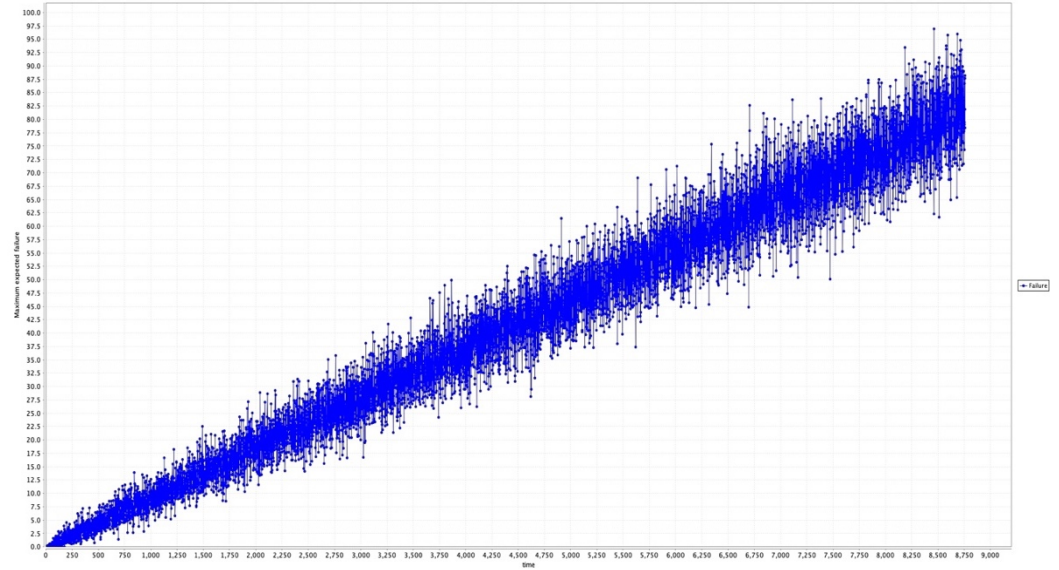

**Figure 8.** Maximum Reward obtained for the Failure Operation mode of the IPR in one year of operation.

In Figure 9, the results of a maximum expected reward experiment for the Fault 1 operation mode of the IPR over a year of operation are presented in a plot. The maximum cumulative reward was 6, which indicates a total of 6 hours in which the device was in a type 1 fault in the simulation.

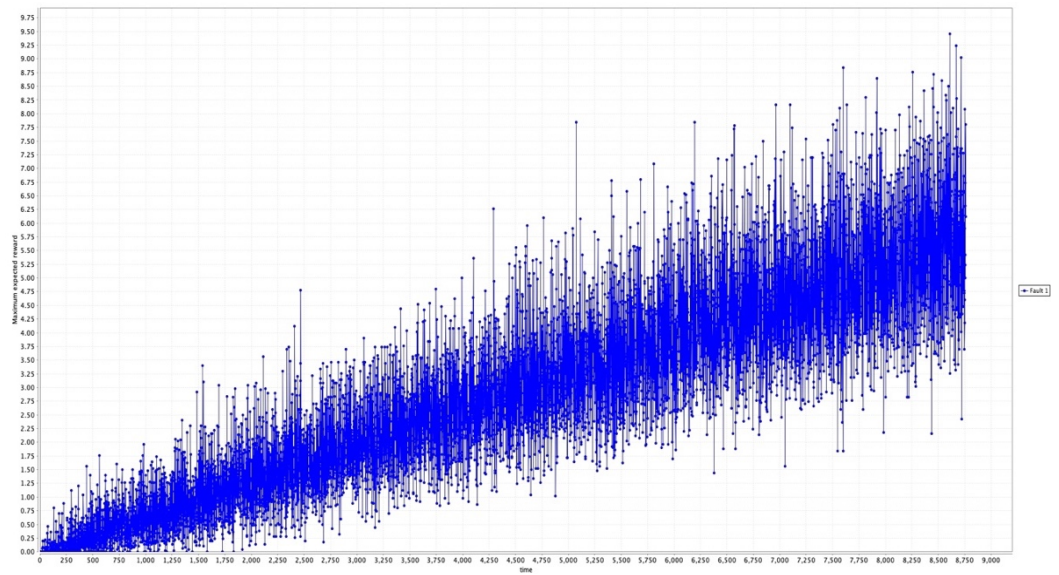

**Figure 9.** Maximum Reward obtained for the Fault 1 Operation mode of the IPR in one year of operation.

Figure 10 presents the result of the maximum cumulative reward for the Fault 2 operation mode of the IPR over a year of operation, and this was found to be 42, reflecting 42 hours in the simulation that the device spent in a type 2 fault.

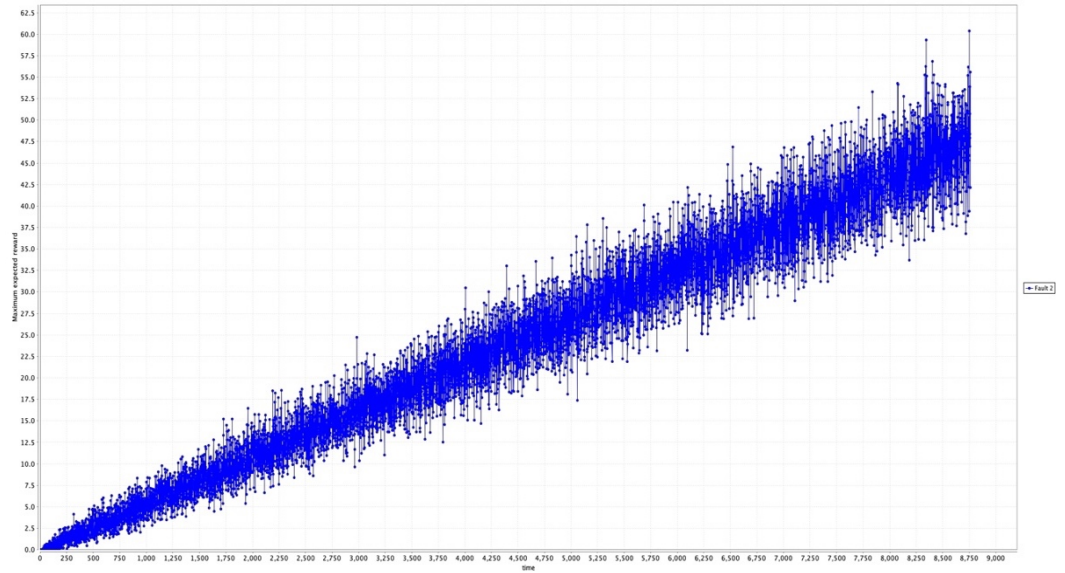

**Figure 10.** Maximum Reward obtained for the Fault 2 Operation mode of the IPR in one year of operation.

These results demonstrate that the PBN model of the IPR can correctly emulate the standard RL cycle, as for every iteration in time, the variables' states are assessed and compared to the failure modes of the device, and a reward is assigned to the actions that the user wants to reinforce. The results of these experiments are exportable on PRISM as a comma-separated values file that can later be used to analyze the model's behavior using statistical packages or machine learning tools.

In these systems, modeled with complex systems tools, learning occurs in the fundamental sense of adaptation to changes; the system adapts in order to survive. Complex systems self-organize into steady states, which are the long-term behavior of the system, and this self-organization in the most basic sense a form of learning (considering learning as a type of adaptation and self-organization as an adaptive mechanism). The evolution of these complex systems can be controlled externally, through interventions [7] so that the system can avoid "unhealthy" states (failures or faults).

In a standard RL model, the agent makes decisions and is connected to its environment through perception and action. Agent and environment interact during a sequence of time steps, and at each interaction step our agent senses the environment for information and determines the state of its "world". Based on this information, the agent takes an action. The information of the state of the environment constitutes the input of our agent, and the action chosen by our agent becomes its output. The actions taken in each step change the state of its environment, and its own state. A time step later, the state transition's value following the action taken is given to our agent by its environment as a reward. The model is an RL agent acting on its environment and receiving a reward that reinforces a certain behavior over others. Rewards were assigned to the IPR RL-Agent acting upon the information on its environment. A reward of '1' was assigned to the state in which all of the IPR components are operating correctly, (Cat. 2), and no rewards for other failure modes. A rewards-based property in PRISM was used to run an experiment in which an IPR is operating continuously for a year, and we obtained the maximum reward assigned to that action. In Figure 4, the RL agent's goal is to remain in a normal operating mode. If instead of positively rewarding the correct operation of the IPR, we rewarded one of its failure modes, such as a Category 1 Failure mode, the resulting experiment would yield a maximum reward as per Figure 9. The maximum rewards obtained are directly correlated with the estimated reliability of the IPR in [1]. Therefore, this reward result is related to the occurrence of the Category 1 failure mode. As a reminder, PRISM does not currently support negative rewards. Validation through experiments that PBN modeled systems can perform the standard RL cycle is an important step towards an automatic way of system control through Machine Learning, where control is not external, but intrinsic to the system.

## 5 Conclusions

In this work, we have studied a smart grid management device, the Intelligent Power Router, assessed the device's reliability, and presented a bioinspired modeling technique that uses Probabilistic Boolean Networks to create simple and logical models that exhibit complex behavior. These models self-organize into constituent Boolean networks with attractor cycles that can describe long-term system behavior. In this case, this behavior is equivalent to the different states in which the device's components can be, and, therefore, to the different failure modes of the device. We used PBNs to model Intelligent Power Routers, a Smart grid component, with Reinforcement Learning, and we studied the model's evolution in time and how they may learn to avoid undesirable states in an autonomous way, increasing their reliability and resilience. We proposed PBNs as a building block with a novel analysis technique for problem solving in Smart grid modeling, through RL. We validated and verified through Model Checking (MC) the viability of this methodology.

We believe that many areas of future work are available for the unanswered questions in this research. Particularly, we have only explored modeling the standard RL cycle, but we infer it is possible to use other RL techniques, such as Q-Learning and Deep Q-Learning, to endow the system with automatic machine-learning based control of its evolution. There is a more fundamental question that can be answered through the study of RL in PBN modelling; that an alternative to Artificial Neural Networks can be achieved through the use of PBNs as the building block for this structure. In order to achieve this, Artificial Neural Network neurons need to be proven equivalent to the set of nodes of a PBN, which have input and output states. The set of input nodes have a set of predictor functions that define the output state (like the threshold function). The learning task would be to change the transition probabilities in order to select a context (constituent BN) that represents the state in which the network has to be (steady state), where states are the goals to be achieved in the system.

## 5 References

1. Irizarry-Rivera, A., Rodríguez-Martínez, M, Vélez, B. Vélez-Reyes, M., Ramírez-Orquín, A.R., O'Neill-Carrillo, E., Cedeño, J. R. (2010). Intelligent Power Routers: Distributed Coordination for Electric Energy Processing Networks. In "Operation and Control of Electric Energy Processing Systems" James Momoh and Lamine Mili (Eds). John Wiley and Sons. Hoboken, New Jersey, pp. 47-85.
2. Gershenson, Carlos (2013). Facing Complexity: Prediction vs. adaptation. In Massip, A. and Bastardas, A., Eds., Complexity Perspectives on Language, Communications and Society, pp. 3-14, Springer-Verlag, Berlin.
3. Rivera Torres, Pedro J., Kanaan Izquierdo, S. (2020). Contributions to Reinforcement Learning through Probabilistic Boolean Networks, a poster presented in the Conference on Complex Systems, held in Thessaloniki, Greece, October.
4. Rivera Torres, Pedro J., Llanes Santiago, O. Fault Detection and Isolation in Smart Grid Devices using Probabilistic Boolean Networks. Computational intelligence in emerging technologies for engineering applications. February, 2020, Springer.
5. Rivera Torres, Pedro J., Silva Neto, Antônio J., Llanes Santiago, O. (2019). Multiple Fault Diagnosis in Manufacturing Processes and Machines using Probabilistic Boolean Networks. In: Advances in Intelligent Systems and Computing, Martínez Álvarez F., Troncoso Lora A., Sáez Muñoz J., Quintián H., Corchado E. (eds), SOCO 2019, Vol. 950, pp. 355-365. Springer.
6. Rivera-Torres, Pedro J., Serrano Mercado, E.I., Llanes Santiago, O., & Anido Rifon, L. (2018a). Modeling Preventive Maintenance of Manufacturing Processes with Probabilistic Boolean Networks with Interventions, Journal of Intelligent Manufacturing, Vol. 29(8), pp. 1941- 1952, Springer.
7. Rivera-Torres, Pedro J., Serrano Mercado, E.I., and Anido Rifon, L. (2018b). Probabilistic Boolean Networks and Model Checking as an Approach for DFMEA for Manufacturing Systems, Journal of Intelligent Manufacturing, Vol. 29(6), pp. 1393-1413, Springer.

8. Rivera-Torres, Pedro J., Serrano Mercado, E.I., & Anido Rifon, L. (2018c). Probabilistic Boolean Network Modeling of an Industrial Machine, Journal of Intelligent Manufacturing, Vol. 29(4) pp. 875-890, Springer.
9. Rivera Torres, Pedro J., Llanes Santiago, O. (2018d). Model-based Fault Diagnosis of Manufacturing Processes and Machines using Probabilistic Boolean Networks. 19th Scientific Conference of Engineering and Architecture, ISP-CUJAE, La Habana, Cuba, November.
10. Rivera Torres, Pedro J. (2017). Contribuciones al modelado teórico de sistemas de fabricación mediante Redes Booleanas Probabilísticas. Doctoral Dissertation, Universidade de Vigo, Spain March.
11. Rivera Torres, Pedro J., Serrano Mercado, E.I. (2016). Probabilistic Boolean Network Modeling as an aid for DFMEA in Manufacturing Systems. 18th Scientific Conference of Engineering and Architecture, ISP-CUJAE, La Habana, Cuba, November.
12. Rivera-Torres, Pedro J., Seguel, Jaime, Rodríguez-Martínez, Manuel, and Irizarry-Rivera, Agustín (2012). Formal Methods for the Design and Analysis of Electrical Power Distribution Systems Endowed with Intelligent Power Routers, Submitted as Final Report for PO 4100307844 – LMCO - 2012, a grant funded by Lockheed-Martin.
13. Kauffman, S. A. (1969b). Metabolic stability and epigenesis in randomly constructed genetic nets. *Journal of Theoretical Biology*, 22, 437–467.
14. Kauffman, S. A. (1969a). Homeostasis and differentiation in random genetic control networks. *Nature*, (224), 177–178.
15. Shmulevich, I., Dougherty, E., & Kim, S. (2002a). Probabilistic Boolean networks: a rule-based uncertainty model for gene regulatory networks. *Bioinformatics*. http://bioinformatics.oxfordjournals.org.ezproxy.library.wisc.edu/content/ 18/2/261.short .
16. Shmulevich, I., Dougherty, E. R., Kim, S., & Zhang, W. (2002b). From Boolean to Probabilistic Boolean Networks as Models of Genetic Regulatory Networks. *Proceedings of the IEEE*, 90, 1778–1792.
17. Shmulevich, I., & Dougherty, E. R. (2007). *Genomic Signal Processing* (1st ed., Vols. 1-1, Vol. 1). Princeton: Princeton University Press.
18. Shmulevich, I., & Dougherty, E. R. (2010). *Probabilistic Boolean Networks: Modeling and Control of Gene Regulatory Networks*. Philadelphia, PA, USA: SIAM.
19. Arnosti, D. N., & Ay, A. (2012). Boolean modeling of gene regulatory networks: Driesch redux. Proceedings of the National Academy of Sciences, 109(45), 18239–18240.
20. Bane, V., Ravanmehr, V., & Krishnan, A. R. (2012). An information theoretic approach to constructing robust Boolean gene regulatory networks. IEEE/ACM Transactions on Computational Biology and Bioinformatics, 9(1), 52-65.
21. Didier, G., & Remy, E. (2012). Relations between gene regulatory networks and cell dynamics in Boolean models. Discrete Applied Mathematics, 160(15), 2147–2157.
22. Ghanbarnejad, F. (2012). Perturbations in Boolean Networks as Model of Gene Regulatory Dynamics (Doctoral Thesis). University of Leipzig, Leipzig, Germany.
23. Chaouiya, C., Ourrad, O., & Lima, R. (2013). Majority Rules with Random Tie-Breaking in Boolean Gene Regulatory Networks. PLoS ONE, 8(7).
24. Ching, W.-K., Chen, X., & Tsing, N.-K. (2009). Generating Probabilistic Boolean Networks from a Prescribed Transition Probability Matrix. IET Systems Biology, 3, 453–464.
25. Kobayashi, K., & Hiraishi, K. (2010). Reachability Analysis of Probabilistic Boolean Networks using Model Checking (pp. 829–832). Presented at the SICE Annual Conference 2010, Proceedings of. http://library.uprm.edu:2055/stamp/stamp.jsp?tp=&arnumber=5604207.
26. Chen, X., Jiang, H., & Ching, W.-K. (2012). On Construction of Sparse Probabilistic Boolean Networks. East Asian Journal on Applied Mathematics. doi:10.4208/eajam.030511.060911a.
27. Gao, Y., Xu, P., Wang, X., & Liu, W. (2013). The complex fluctuations of Probabilistic Boolean Networks. BioSystems, 114(1), 78–84.
28. Trairatphisan, P., Mizera, A., Pang, J., Tantar, A. A., Schneider, J., & Sauter, T. (2013). Recent development and biomedical applications of Probabilistic Boolean Networks. Cell Communication and Signaling, 11, 46.
29. Chen, H., & Sun, J. (2014). Stability and Stabilisation of context-sensitive Probabilistic Boolean Networks. IET Control Theory & Applications, 8(17), 2115–2121.
30. Gershenson, Carlos. (2004). Introduction to Random Boolean Networks. In Bedau, M., Hutton, T., Kumar, S., and Suzuki, H., Eds. Workshop and Tutorial Proceedings, Ninth International Conference on the Simulation and Synthesis of Living Systems (Alife IX). pp. 160-173, Boston MA.
31. Gershenson, C. (2012). Guiding the self-organization of random Boolean networks. Theory in Biosciences, 131(3): 181-191.

32. Sutton, R. S. and A. G. Barto (1998a). Macro-Actions in Reinforcement Learning: An Empirical Analysis' by Amy McGovern and Richard S. Sutton.
33. Sutton, R. S. and A. G. Barto (1998b). Reinforcement Learning: An introduction, vol. 1, Cambridge, Massachusetts: MIT Press.
34. McCallum, A. K. (1996). Reinforcement Learning with Selective Perception and Hidden State. Ph.D. Dissertation, University of Rochester.
35. Rummery, G. A. (1995). Problem Solving with Reinforcement Learning. Ph.D. Dissertation, Cambridge University, Cambridge, UK.
36. Rivera Torres, P. J. and Llanes Santiago, O. (2020). Fault Detection and Isolation in Smart Grid Devices Using Probabilistic Boolean Networks. In: Llanes Santiago, O., Cruz Corona, C., Silva Neto, A.J. and Verdegay, J.L., (Eds). Computational Intelligence in Emerging Technologies for Engineering Applications. Studies in Computational Intelligence, Vol. 872 Springer-Nature, Switzerland.
37. Kwiatkowska, M. Z., Norman, G., & Parker, D. (2011). PRISM 4.0: Verification of Probabilistic Real-Time Systems. In *Lecture Notes in Computer Science.* 6806, 585–591.
38. Mendonça, L.F., Sousa, J.M., Sá da Costa, J.M. (2009). An architecture for fault detection and isolation based on fuzzy methods. Expert Systems with Applications. (36), 1092-1104.